\documentclass[]{bytedance_seed}

\usepackage[toc,page,header]{appendix}

\usepackage{natbib}
\usepackage{hyperref}
\usepackage{url}
\usepackage{xspace}
\usepackage{amsfonts}
\usepackage{amsmath}
\usepackage{algorithm}
\usepackage{algorithmic}
\usepackage{graphicx}
\usepackage{booktabs}
\usepackage{multirow}
\usepackage{subcaption}
\usepackage{longtable}
\usepackage[table]{xcolor}
\usepackage{enumitem}
\usepackage{tabularx}
\usepackage{threeparttable}
\usepackage[most]{tcolorbox}
\usepackage{siunitx}

\newcommand{\ie}{\emph{i.e.}\xspace}

\newcommand{\ignore}[1]{}

\definecolor{colorA}{rgb}{ .882,  .949,  .89}
\definecolor{colorB}{rgb}{ .922,  .89,  .988}
\definecolor{colorC}{rgb}{ 1.000, .976, .831}

\title{AgenticGen: Reward-Guided Agentic Video Generation for Advertising}
\author[\dagger\ddagger1]{Xingyuan Bu}
\author[1]{Chengru Song}
\author[1]{Hao Zhou}
\author[1]{Tao Zhou}
\author[1]{Dong Li}
\author[1]{Wei Li}
\author[1]{Shilong Li}
\author[1,2]{Hao Shi}
\author[1]{Yongxin Guo}
\author[1,3]{Donghao Zhou}
\author[1]{Qiangpeng Yang}
\author[1]{Shilei Wen}

\affiliation[1]{ByteDance}
\affiliation[2]{Tsinghua University}
\affiliation[3]{The Chinese University of Hong Kong}

\contribution[\dagger]{Corresponding Author}
\contribution[\ddagger]{Project Leader}

\abstract{

\begin{abstract}

Advertising video generation is not only a video synthesis task, but also a product-conditioned reasoning problem whose success is measured by online business metrics. Recent video foundation models can generate realistic clips from multimodal conditions, yet they do not optimize how a product should be transformed into an effective advertisement or how future generation should be improved from online business feedback. To close this loop, we propose \textsc{AgenticGen}, a reward-guided agentic framework that decomposes advertising video generation into two trainable reasoning stages, strategy selection and draft generation, thereby exposing optimization targets that online business feedback can supervise. \textsc{AgenticGen} learns a performance-based reward from accumulated online feedback and a complementary rubric-based reward aligned with human quality standards, then uses them to supervise policy optimization. DPO first moves the agentic policies toward online preferences, and GRPO further refines both stages with process and outcome rewards. Offline experiments validate the reward models and successive policy optimization. Online A/B experiments in the TikTok advertising system show that \textsc{AgenticGen} after DPO and GRPO improves CTR by 2.72\%, CVR by 2.63\%, and Advv by 9.61\% over the SFT baseline.
\end{abstract}

}

\date{\today}
\correspondence{Xingyuan Bu at \email{xingyuanbu@gmail.com}, \email{xiyu.xy@bytedance.com}}

\begin{document}
\maketitle


\vspace{-1mm}
\section{Introduction}
\vspace{-1mm}

Advertising video generation is a high-value creative task whose success is ultimately determined not only by visual quality, but also by its impact on online business metrics such as click and conversion. Recent video foundation models~\cite{seedance2026seedance,openai2025sora2,veo3,wan2025wan}, such as Seedance 2.0 and Sora, have substantially advanced multimodal video generation from text, image, audio, and video conditions. Yet effective advertising demands more than generating a realistic clip from a prompt or reference. It requires deciding how heterogeneous product materials should be transformed into a platform-native video that communicates product value and performs effectively after delivery. The task is therefore better formulated as a product-conditioned reasoning problem, in which product materials, prior advertising experience, and expected online performance jointly affect generation decisions.

Advertising delivery, in turn, provides a natural feedback loop for improving these decisions: generated videos are served to users, and their online commercial performance is measured after delivery. In performance advertising, where both advertisers and platforms optimize measurable user response, this cycle turns advertising video generation into a reward-guided decision problem in which accumulated online feedback should improve the decisions that shape subsequent videos. However, converting this loop into a learning system requires more than measuring delivery metrics. It requires learnable reasoning processes that expose meaningful optimization targets, reliable reward signals constructed from delivery data under advertising system bias, and an efficient mechanism for translating accumulated feedback into better generation decisions. The central challenge is therefore to establish a learnable framework that closes the loop from online business feedback to advertising video generation.

To address this challenge, we propose \textsc{AgenticGen}, a reward-guided agentic framework for advertising video generation. \textsc{AgenticGen} uses a strong reasoning Vision-Language Model (VLM) to drive the agent framework and decomposes advertising video generation into two reasoning stages: strategy selection and draft generation. In the strategy selection stage, the agent selects among a set of creative strategies, including asset editing, reference-guided generation, and cross-asset remixing. In the draft generation stage, it turns the selected strategy group into an executable draft with tool calls and generation instructions, then invokes video generation models such as Seedance 2.0~\cite{seedance2026seedance} together with internal rendering tools such as CapCut~\cite{CapCut} to produce the final advertising video. The resulting reasoning processes give online business feedback concrete optimization targets at both stages.

Learning these reasoning processes from online business feedback is necessary but cannot be treated as a direct online reinforcement learning (RL) problem. Reliable feedback is time-consuming because each generated video must be delivered to users and observed over a sufficiently long measurement period before a stable reward can be estimated. The signal is also noisy because advertising system bias and stochastic user behavior can obscure the contribution of the generated advertising video. We therefore first deploy the SFT policies to an impression-balanced delivery pipeline to collect reliable online performance data. From the accumulated feedback, we train a Bradley-Terry (BT) performance-based reward model. In parallel, we train a rubric-based reward model on human-labeled data to capture video defects and clickbait~\cite{wang2021clicks, immorlica2024clickbait}. Before on-policy optimization, DPO~\cite{rafailov2023direct} warms up the strategy selection and draft generation policies using impression-balanced pairwise preference data. Building on this warm start, GRPO~\cite{shao2024deepseekmath} performs on-policy RL with process rewards derived from business preference priors for strategy selection and outcome rewards provided by the two reward models for draft generation.

To evaluate the effectiveness of \textsc{AgenticGen}, we conduct comprehensive offline experiments using performance data and human-labeled data, followed by online A/B experiments in the TikTok advertising system. After DPO and GRPO optimization, \textsc{AgenticGen} improves CTR by 2.72\%, CVR by 2.63\%, and Advv by 9.61\% over the SFT baseline. 

Our main contributions are summarized as follows:
\begin{itemize}[leftmargin=4mm]
    \item To the best of our knowledge, we are the first to formulate advertising video generation as a reward-guided agentic framework. We propose \textsc{AgenticGen}, which decomposes advertising video generation into strategy selection and draft generation so that both reasoning stages can be explicitly optimized with online business feedback.
    \item We construct complementary reward signals for advertising video generation. A performance-based reward model is trained from accumulated online feedback under matched product contexts, while a rubric-based reward model applies domain-specific quality criteria to capture clickbait and severe video defects.
    \item We develop a two-phase reward-guided optimization pipeline for \textsc{AgenticGen}. The DPO training phase learns business performance preferences from the impression-balanced delivery data, while the GRPO training phase performs on-policy optimization by combining process rewards for strategy selection and outcome rewards for draft generation.
    \item We validate \textsc{AgenticGen} through offline evaluation and online A/B experiments in the TikTok advertising system. The optimized agentic framework improves over the SFT baseline, lifting CTR by 2.72\%, CVR by 2.63\%, and Advv by 9.61\%.
\end{itemize}

\section{Related Work}

\noindent \textbf{Agentic Video Generation.}
Video foundation models have rapidly advanced from text-conditioned synthesis to high-fidelity multimodal generation with stronger controllability~\cite{seedance2026seedance,openai2025sora2,veo3,wan2025wan}. Nevertheless, these models remain reactive generators. They do not autonomously decompose creative goals, select references and tools, or revise outputs over a long generation horizon.

Agentic generation addresses this gap by placing foundation models inside planning, execution, and evaluation loops. Early systems use LLM planning and specialized agents to coordinate multi-scene video creation and storytelling~\cite{lin2023videodirectorgpt,yuan2024mora,hu2024storyagent,huang2026vimax,hu2026cameraartist}. Recent video agents extend this paradigm with self-refinement and reusable skills~\cite{long2026vista,song2026vqqa,song2026codirector,wei2026videoweaver,wang2026tempact}, while parallel image agents use tool-augmented multimodal reasoning to improve image synthesis~\cite{jiang2026genagent,chen2026unify,chen2026genevolve}. Their optimization objectives remain centered on offline perceptual quality. \textsc{AgenticGen} instead optimizes video generation with rewards derived from online business feedback at industrial scale.

\noindent \textbf{Reward-guided Generation.}
Policy optimization from feedback includes both off-policy and on-policy families. Off-policy methods optimize policies using fixed preference data, with representative objectives including DPO~\cite{rafailov2023direct}, IPO~\cite{azar2024ipo}, and KTO~\cite{ethayarajh2024kto}. On-policy methods sample from the current policy using PPO~\cite{schulman2017ppo}, REINFORCE~\cite{ahmadian2024back}, GRPO~\cite{shao2024deepseekmath}, or Pref-GRPO~\cite{wang2025pref}.

These optimization methods require supervision that specifies which outputs are preferred. Existing work obtains such supervision from human feedback or evaluation rubric and operationalizes it through Bradley-Terry models~\cite{bradley1952rank}, or reasoning-based evaluators~\cite{wang2025unified,he2024videoscore,wang2026think,huang2024vbench}, and rubric-based judges~\cite{liu2026cdrrm}. The resulting rewards have improved open-ended textual~\cite{jia2025writing} and visual generation~\cite{li2024t2v,liu2025improving}.

Closer to our setting, some industrial generation systems obtain supervision from online business feedback rather than human annotations or offline evaluators. Meta's AdLlama~\cite{jiang2025improving} uses historical CTR as performance feedback. GenCO~\cite{gu2025generative} assigns click-derived rewards to creative combinations. Recent methods also use CTR-derived supervision to generate or refine ad text and images~\cite{wang2026relate,mishra2020learning,chenyanda2025ctr,chenxingye2025ctr,wei2022creater}. Unlike these text- or image-oriented systems, \textsc{AgenticGen} uses online business feedback to optimize the agentic video generation, complemented by a rubric-based reward aligned with human quality standards.

\section{The AgenticGen Framework}
\label{sec:agenticgen-framework}


\textsc{AgenticGen} formulates advertising video generation as a two-stage reasoning pipeline over product materials. It first selects a feasible strategy group conditioned on product, and then generates an executable draft for video generation models and rendering tools. This design exposes trainable optimization targets for online business feedback.

\begin{figure*}[t]
    \centering
    \includegraphics[width=0.99\textwidth]{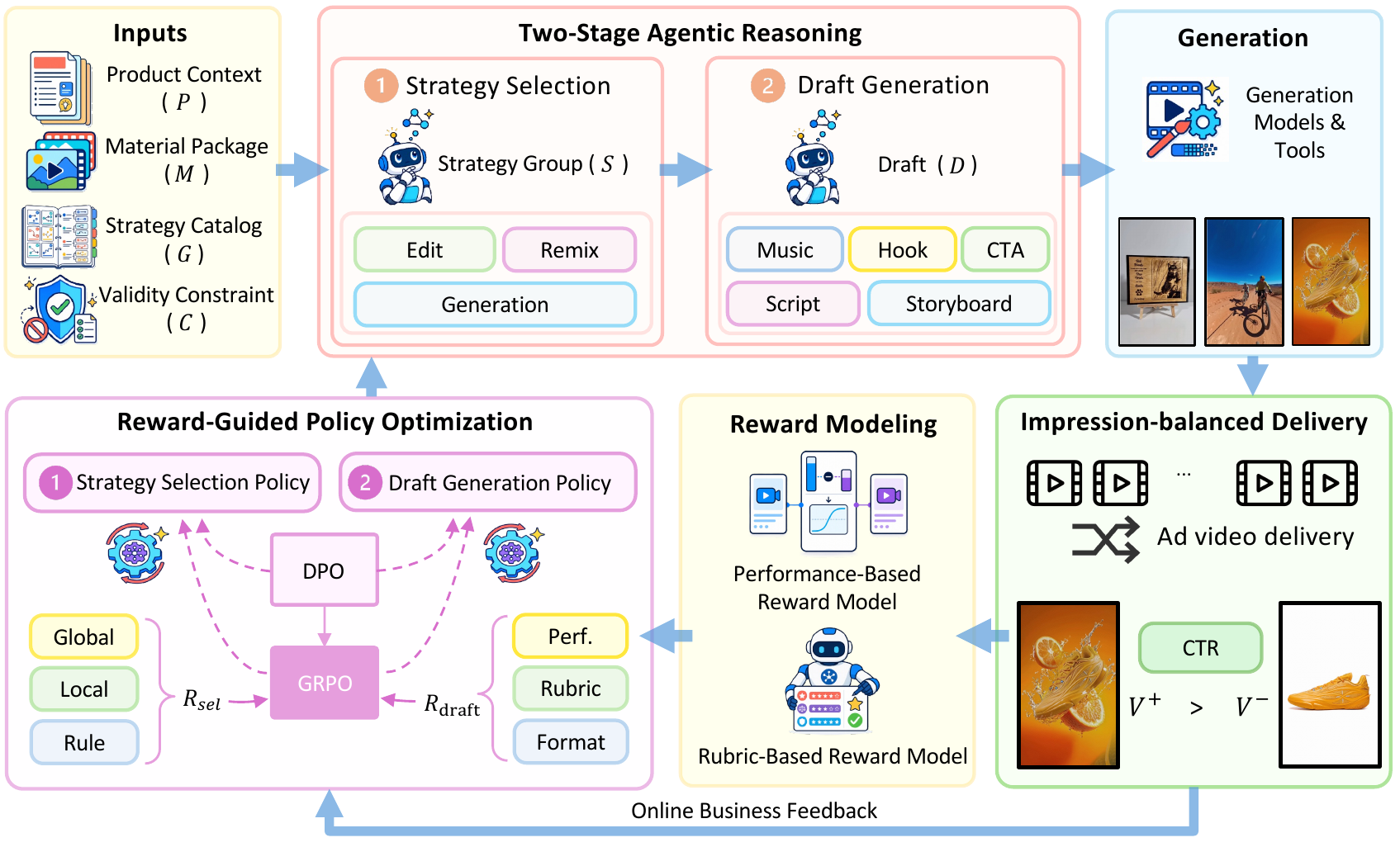}
    \caption{Overview of the \textsc{AgenticGen} framework and its online optimization loop. Given product context $P$, material package $\mathcal{M}$, strategy catalog $\mathcal{G}$, and validity constraints $\mathcal{C}$, a reasoning VLM first selects a feasible strategy group $S$ and then converts it into an executable draft $D$ for video generation models and tools. The generated videos are served through impression-balanced delivery to obtain comparable CTR feedback. This feedback trains the performance-based reward model and constructs preference data for DPO, while expert quality standards supervise the rubric-based reward model. After DPO, GRPO uses process rewards for strategy selection and outcome rewards for draft generation, and alternately optimizes the two policies.}
    \label{fig:agenticgen-overview}
\end{figure*}

Figure~\ref{fig:agenticgen-overview} summarizes the complete framework and its optimization loop. The remainder of this section describes the strategy selection and draft generation pipeline, followed by SFT initialization and online feedback collection. Section~\ref{sec:reward-optimization} then presents the two reward models and the DPO and GRPO optimization phases.

\subsection{Strategy Selection and Draft Generation Pipeline}
\label{sec:agentic-production}

Rather than generating the final video in one step, \textsc{AgenticGen} decomposes the process into strategy selection and draft generation.

\noindent \textbf{Strategy Selection.}
Strategy selection takes a product context $P$ and a material package $\mathcal{M}$ as product-specific inputs. Here, $P$ contains product information such as its name, description, and selling points, while $\mathcal{M}$ combines source video and image assets with their historical performance and parsed multimodal features. It then selects a strategy group $S$ from a predefined strategy catalog $\mathcal{G}=\{s_i\}_{i=1}^{|\mathcal{G}|}$, subject to validity constraints $\mathcal{C}$:
\begin{equation}
\label{eq:strategy-selection}
S=\mathcal{F}_{\mathrm{sel}}(P,\mathcal{M}; \mathcal{G},\mathcal{C}).
\end{equation}
Unlike general video generation, advertising video generation operates with existing high-quality advertiser materials and explicit delivery objectives. To reuse these materials while complementing them with newly generated video clips, internal advertising experts define more than a dozen strategies based on TikTok video creative priors. These strategies form $\mathcal{G}$ and can be roughly grouped into three public families: asset editing, reference-guided generation, and cross-asset remixing. See Appendix~\ref{app:strategy-examples} for strategy definitions.

The selection function $\mathcal{F}_{\mathrm{sel}}$ is implemented by a strong reasoning VLM that consumes all inputs and chooses suitable strategies from $\mathcal{G}$. During chain-of-thought (CoT) reasoning, the VLM considers product information, parsed multimodal material features, historical material performance signals, and source video quality to balance creative diversity with expected advertising value. For example, if an existing product video already has strong online performance and usable visual quality, the VLM may select lightweight editing strategies, such as replacing music with a stronger track or strengthening the call to action (CTA). If the original material is insufficient or the source video quality is poor, the VLM may select reference-guided generation to create new material and then remix it with other materials and insert marketing visual elements through editing strategies.

The VLM also reasons under the validity constraints in $\mathcal{C}$ while choosing strategies. For example, remixing is invalid when there are not enough eligible clips with sufficient duration, and background music replacement is invalid when the source material contains singing. The output of this stage is a selected strategy group $S=\{s_k\}_{k=1}^{K}$, where each $s_k\in S$ is feasible under the current product context and material package.

\noindent \textbf{Draft Generation.}
Draft generation converts the selected strategy group into concrete executable content and parameters. We write this stage as
\begin{equation}
\label{eq:draft-generation}
D=\mathcal{F}_{\mathrm{draft}}(P,\mathcal{M},S).
\end{equation}
Draft generation also utilizes the reasoning VLM to turn each selected strategy into an executable structured draft. For each $s_k\in S$, the VLM first plans over the product context, source materials, and selected strategies, and then writes the corresponding draft fields $d_k\in D$. The draft fields vary across strategies. For example, lightweight editing strategies may specify choices such as music, stickers, overlays, and short selling points and CTA content. Remix strategies may rewrite the storyline and select and order source clips. Reference-guided generation may retrieve and interpret top-performing videos from the advertising traffic pool and produce a storyboard script that imitates highly effective narrative patterns while adapting them to the current product.

After the draft format is validated, the execution layer maps $D$ to the final advertising video:
\begin{equation}
\label{eq:generation-execution}
V=\mathcal{F}_{\mathrm{gen}}(D).
\end{equation}
The $\mathcal{F}_{\mathrm{gen}}$ is not an additional reasoning stage. It executes the generation instructions, editing, and rendering operations specified in $D$ through video generation models such as Seedance 2.0~\cite{seedance2026seedance} and internal rendering tools such as CapCut~\cite{CapCut}, making $D$ the interface between the agentic reasoning process and the final video artifact.

\subsection{SFT Initialization and Online Feedback Collection}
\label{sec:online-data}

\noindent \textbf{SFT Initialization.}
For the initial deployment of the agentic framework, we perform supervised fine-tuning (SFT) on Qwen3-VL-8B-Thinking~\cite{bai2025qwen3vl} using data generated by a larger reasoning model. These SFT trajectories cover the reasoning processes and structured outputs for both strategy selection and draft generation. Deploying the smaller SFT model reduces serving costs compared with using the larger teacher model directly. The deployed SFT model also serves as an online data-collection policy, since its generated videos can be delivered online and their online business performance can be joined with the corresponding logged trajectories for the later DPO and GRPO training phases.

\noindent \textbf{Impression-Balanced Delivery Pipeline.}
Directly delivering generated videos through the standard delivery pipeline would place them in the same material pool as advertisers' original videos and cause them to go through recall and ranking. The resulting online business metrics would therefore be confounded by cold-start effects, traffic composition, and other sources of uneven delivery opportunity. To obtain cleaner online preference data, we develop an impression-balanced delivery pipeline.

The pipeline reserves a dedicated pool of online traffic. For each product, a group of $N=12$ videos generated by \textsc{AgenticGen} is attached to the original product, routed directly to impression serving, and shown to users through random choices. This design bypasses the advertising system's upstream recall and ranking stages and balances impressions among the videos within the same delivery context. Under this protocol, differences in online business metrics are more directly attributable to relative video quality than to uneven delivery opportunity. We primarily use click-through rate (CTR) as the online performance metric. To ensure that the collected feedback is reliable, we consider a delivered video valid for training only after it receives at least 1000 impressions.

For each delivered video, we log the complete \textsc{AgenticGen} trajectory corresponding to Eqs.~\ref{eq:strategy-selection}--\ref{eq:generation-execution}. The trajectory contains the product context, material package, selected strategy group, structured draft, rendered video, and corresponding online business feedback. These trajectories are used for both reward modeling and reinforcement learning.

\section{Reward Modeling and Policy Optimization}
\label{sec:reward-optimization}

The feedback collected through the deployed SFT model supports both reward learning and policy optimization. We first train a performance-based reward model from impression-balanced online feedback and a rubric-based reward model from domain-specific human quality standards. We then use impression-balanced preference data to warm up the strategy selection and draft generation policies with DPO, followed by on-policy GRPO using process rewards for strategy selection and outcome rewards for draft generation.

\subsection{Performance-Based Reward Model}

The performance-based reward model learns video preferences from online business feedback. Direct pointwise regression on absolute metrics such as CTR is suboptimal because observed values conflate intrinsic video appeal with traffic composition and advertising system bias, producing noisy and unstable supervision. We therefore adopt a Bradley-Terry pairwise comparison objective~\cite{bradley1952rank,liu2025improving}, which compares videos within matched delivery contexts, mitigates confounders, and learns more reliable relative preferences.

Using the impression-balanced data, we construct pairwise training examples within each group of $N$ videos. Once sufficient online business feedback has accumulated, we identify the highest and lowest performing videos in the same group and use them as the positive and negative examples, respectively. Since the compared candidates share the same product context and delivery setting while receiving balanced impression opportunities, these within-group comparisons provide cleaner relative preference labels for reward modeling.

Advertising video performance depends on high-level creative factors such as narrative appeal and marketing salience. Relying only on raw video frames makes these semantics difficult to learn from online feedback. We therefore expose four feature groups to the reward model:
\begin{itemize}[leftmargin=4mm]
    \item video frames, for enabling the model to assess visual quality, product visibility, scene appeal, and clip transitions.
    \item audio and speech features, including the audio track itself, background music, and ASR descriptions, for capturing the video's overall emotion, rhythm, and spoken content.
    \item dense captions and structured storylines, which turn frame-level events and video-level story into textual semantics for easier understanding of the video narrative logic.
    \item ad-specific features, including hooks, subtitles, stickers, selling-points, product names, brand names, CTA cues, and language, for more direct expression of key commercial information.
\end{itemize}
These features are then encoded by the reward model, so the model can compare not only what appears in a video, but also how it presents the product.

Let $\mathcal{D}_{R}$ denote the pairwise training data. Each example $(P,V^+,V^-)\in \mathcal{D}_{R}$ contains a product context $P$ and two videos compared within that context: a high performing video $V^+$ and a low performing video $V^-$. The performance-based reward model $r_\theta$ outputs scalar scores $r_\theta(P,V^+)$ and $r_\theta(P,V^-)$, and is trained with
\begin{equation}
\label{eq:bt-reward}
\begin{aligned}
\mathcal{L}_{R}
&= -\mathbb{E}_{\mathcal{D}_{R}}
\left[\log \sigma\left(r_\theta(P,V^+)-r_\theta(P,V^-)\right)\right],
\end{aligned}
\end{equation}
where $\sigma$ is the sigmoid function. The resulting scalar scoring provides dense rewards for draft generation GRPO.

\subsection{Rubric-Based Reward Model}

Online business feedback is the primary signal for commercial value, but it does not fully specify whether an individual video satisfies human quality standards. A generated video may receive favorable online performance feedback while still containing quality issues such as visual glitches or clickbait~\cite{wang2021clicks, immorlica2024clickbait}. To complement the performance-based reward model, we train a rubric-based reward model that judges video quality according to TikTok domain-specific human quality standards. We first ask internal advertising experts to summarize these standards. The resulting standards cover four content dimensions: video script, video content, video decoration, and audio content. Within each content dimension, they further consider five quality aspects: platform-native style, attention grabbing strength, logical coherence, product consistency, and aesthetic harmony. We convert these expert summarized standards into rubric prompts.

As discussed in the Performance-Based Reward Model section, learning targets can drift when videos are judged without a comparable context. The same issue applies to rubric scoring~\cite{liu2025improving}. We therefore use a pairwise prompt that contains a reference video $V_r$, a candidate video $V_c$, their respective feature groups described above, the product context $P$, and the rubric. We train a Qwen2.5-Omni-7B~\cite{xu2025qwen25omni} to reason over the four content dimensions and five quality aspects and output whether the candidate video is better than the reference video.

We collect the rubric reward training data through human annotation. Each video pair is labeled by three annotators. We observe that when annotators disagree substantially, the relative quality of the videos is often intrinsically ambiguous and affected by individual preferences. When all three annotators give the same label, the relative quality judgment is more reliable. We therefore retain only these fully agreed pairs, which provide the model with a more stable learning signal.

\subsection{DPO from Impression-balanced Data}

DPO~\cite{rafailov2023direct} serves as the off-policy training phase that moves the SFT agentic policies toward business preferences. This warm-up reduces the burden on the subsequent on-policy GRPO training phase. Since \textsc{AgenticGen} contains strategy selection and draft generation stages, we collect separate off-policy training data for the two stages by fixing one stage and collecting online preference data for the other.

For strategy selection, the optimized output is the selected strategy group $S$ in Eq.~\ref{eq:strategy-selection}. Given the same input, the selection policy samples $N$ strategy groups $\{S_i\}_{i=1}^{N}$, while the draft policy is fixed with temperature 0 to generate $D_i$ and final video $V_i$ for each $S_i$. The $N$ videos are served together through the impression-balanced delivery pipeline. After enough online feedback is accumulated, the strategy group with the highest online business feedback is labeled $S^+$, and the group with the lowest feedback is labeled $S^-$. This yields $\mathcal{D}_{\mathrm{DPO}}^{\mathrm{sel}}$, where the main variation comes from the strategy selection stage. For draft generation, the selected strategy group $S$ is fixed, and the draft policy samples $\{D_i\}_{i=1}^{N}$ under the same input $(P,\mathcal{M},S)$ in Eq.~\ref{eq:draft-generation}. The resulting videos are delivered and labeled by the same highest-lowest way, yielding $\mathcal{D}_{\mathrm{DPO}}^{\mathrm{draft}}$ with preference pairs $(D^+,D^-)$, where the main variation comes from the draft generation stage.

Let $\pi_\psi$ be the trainable policy and $\pi_{\mathrm{ref}}$ be the SFT reference policy. For either reasoning stage, let $X$ denote the input and $Y^+,Y^-$ denote the preferred and dispreferred policy outputs. We optimize the standard DPO objective:
\begin{equation}
\label{eq:dpo-objective}
\begin{aligned}
\mathcal{L}_{\mathrm{DPO}}
= -\mathbb{E}_{\mathcal{D}_{\mathrm{DPO}}}
\Bigg[\log \sigma\Bigg(
\beta \log \frac{\pi_\psi(Y^+ \mid X)}
{\pi_{\mathrm{ref}}(Y^+ \mid X)}
-\beta \log \frac{\pi_\psi(Y^- \mid X)}
{\pi_{\mathrm{ref}}(Y^- \mid X)}
\Bigg)\Bigg],
\end{aligned}
\end{equation}
where $\sigma$ is the sigmoid function and $\beta$ controls KL regularization. We alternately optimize the two stages by instantiating $(X,Y)$ with the corresponding input and output in Eq.~\ref{eq:strategy-selection} or Eq.~\ref{eq:draft-generation}, using $\mathcal{D}_{\mathrm{DPO}}^{\mathrm{sel}}$ for strategy selection and $\mathcal{D}_{\mathrm{DPO}}^{\mathrm{draft}}$ for draft generation.

To further stabilize DPO training by maintaining valid format for generation, we add a negative log-likelihood (NLL) term with weight $\lambda_{\mathrm{nll}}=0.2$ following the practice used in~\cite{grattafiori2024llama,pang2024iterative}. The final training loss is
\begin{equation}
\label{eq:dpo-nll}
\mathcal{L}_{\mathrm{DPO}}^{\mathrm{reg}}
= \mathcal{L}_{\mathrm{DPO}}
+ \lambda_{\mathrm{nll}}
\mathbb{E}_{\mathcal{D}_{\mathrm{DPO}}}
\big[-\log \pi_\psi(Y^+\mid X)\big].
\end{equation}

\subsection{GRPO with Process and Outcome Rewards}

GRPO~\cite{shao2024deepseekmath} is the on-policy training phase that further improves \textsc{AgenticGen} after DPO. Unlike DPO, which learns from fixed preference pairs, GRPO samples new response groups from the current policy, optimizes the policy under its own rollout distribution, and can incorporate multiple verifiable and model-based reward signals during policy updates, making it better suited for the final optimization stage after the DPO warm-up. Since \textsc{AgenticGen} has a long agentic pipeline, a final video reward alone gives sparse supervision to the intermediate reasoning process. We therefore use separate reward modeling for strategy selection and draft generation, so process rewards guide selected strategies while final outcome rewards guide drafts. For GRPO optimization, we also use training data collected from the impression-balanced delivery pipeline because it provides fair comparisons within each sampled group under balanced impressions.

\noindent \textbf{Strategy Selection Reward Modeling.}
At the strategy selection stage, we convert online performance feedback into statistical process rewards and combine them with a rule-based indicator function. Specifically, each sampled strategy group is evaluated with two rewards and one indicator. 
\begin{itemize}[leftmargin=4mm]
    \item Global reward $R_{\mathrm{glob}}$: This reward estimates global strategy priors from impression-balanced logs. We attach each strategy group's feedback value to every member strategy $s_k$, average the assigned feedback for each strategy across products and campaigns, and score a sampled strategy group by the mean of its strategy scores. This aggregation captures strategy performance trends across products and campaigns. See Appendix~\ref{app:selection-reward-details}.
    \item Local reward $R_{\mathrm{loc}}$: This reward measures how well a sampled group matches the best strategy group under the current product context. Since the GRPO training data comes from the impression-balanced delivery pipeline, we can treat the winning strategy group $S^+$ among $\{S_i\}_{i=1}^{N}$ as an approximate best strategy group and reward each rollout sample by their overlap. This construction captures product-level preference. See Appendix~\ref{app:selection-reward-details}.
    \item Rule-based indicator $\mathbb{I}_{\mathrm{sel}}$: This indicator verifies the constraints in $\mathcal{C}$ from Eq.~\ref{eq:strategy-selection} and the output format of $S$. If the check fails, $\mathbb{I}_{\mathrm{sel}}=0$ and the total selection reward is set to zero, preventing invalid strategy groups from entering the agentic pipeline.
\end{itemize}
The global and local rewards have different ranges, so we min-max normalize them into $\widetilde{R}_{\mathrm{glob}}$ and $\widetilde{R}_{\mathrm{loc}}$. The resulting strategy reward is
\begin{equation}
\label{eq:selection-grpo-reward}
R_{\mathrm{sel}} =
\begin{cases}
\lambda_{\mathrm{sel}} \widetilde{R}_{\mathrm{glob}}
+ (1-\lambda_{\mathrm{sel}})\widetilde{R}_{\mathrm{loc}}, & \mathbb{I}_{\mathrm{sel}}=1,\\
0, & \mathbb{I}_{\mathrm{sel}}=0.
\end{cases}
\end{equation}
Here, $\lambda_{\mathrm{sel}}$ controls the trade-off between global and local strategy rewards. Thus, valid strategy groups are optimized toward both global strategy priors and product-level winners, while invalid strategy groups receive zero reward.

\noindent \textbf{Draft Generation Reward Modeling.}
For the draft generation stage, rewards are assigned after video generation because the draft must be executed before reward model scoring. We use two model-based reward signals and one indicator function:
\begin{itemize}[leftmargin=4mm]
    \item Performance reward $R_{\mathrm{perf}}$: This reward scores each generated video with the performance-based reward model and provides a dense preference signal aligned with online business feedback.
    \item Rubric reward $R_{\mathrm{rub}}$: This reward estimates video quality through pairwise rubric judgments. Following Pref-GRPO~\cite{wang2025pref}, we compute each generated video's average win rate against the other videos in the same rollout group.
    \item Format indicator $\mathbb{I}_{\mathrm{draft}}$: This indicator verifies the structured draft format and executable generation calls. If the check fails, $\mathbb{I}_{\mathrm{draft}}=0$ and the total draft reward is set to zero, preventing malformed drafts from causing failures during downstream video generation or rendering.
\end{itemize}
We also min-max normalize the performance and rubric rewards into $\widetilde{R}_{\mathrm{perf}}$ and $\widetilde{R}_{\mathrm{rub}}$. The final draft reward is
\begin{equation}
\label{eq:draft-fusion-reward}
R_{\mathrm{draft}} =
\begin{cases}
\lambda_{\mathrm{draft}} \widetilde{R}_{\mathrm{perf}}
+ (1-\lambda_{\mathrm{draft}})\widetilde{R}_{\mathrm{rub}},
& \mathbb{I}_{\mathrm{draft}}=1,\\
0, & \mathbb{I}_{\mathrm{draft}}=0.
\end{cases}
\end{equation}
Here, $\lambda_{\mathrm{draft}}$ controls the relative contribution of preference and rubric quality rewards.

For either stage, let $X$ denote the stage input, let $\{Y_i\}_{i=1}^{G}$ be the sampled outputs from the rollout policy, and let $\{R_i\}_{i=1}^{G}$ be their final rewards after applying the indicator functions. We compute group-relative advantages as
\begin{equation}
\label{eq:grpo-advantage}
A_i =
\frac{R_i-\operatorname{mean}(\{R_1,\ldots,R_G\})}
{\operatorname{std}(\{R_1,\ldots,R_G\})}.
\end{equation}
Let $\pi_{\psi_{\mathrm{old}}}$ be the rollout policy, $\pi_\psi$ be the updated policy, and $\mathcal{D}_{\mathrm{GRPO}}$ denote the on-policy rollout data for the current update. With $r_i=\pi_\psi(Y_i \mid X)/\pi_{\psi_{\mathrm{old}}}(Y_i \mid X)$, we maximize the GRPO objective:
\begin{equation}
\label{eq:grpo-objective}
\begin{aligned}
J_{\mathrm{GRPO}}(\psi)
= \mathbb{E}_{\mathcal{D}_{\mathrm{GRPO}}}\Bigg[
\frac{1}{G}\sum_{i=1}^{G}
\min\Big(
&r_i A_i,
\operatorname{clip}(r_i,1-\epsilon,1+\epsilon)A_i
\Big)\\
&\qquad - \beta
D_{\mathrm{KL}}(\pi_\psi\|\pi_{\mathrm{ref}})
\Bigg],
\end{aligned}
\end{equation}
where $\epsilon$ and $\beta$ are hyperparameters and $D_{\mathrm{KL}}$ denotes the KL divergence. As in the DPO training phase, we alternately optimize the GRPO strategy selection and draft generation policies using the corresponding data and rewards.

\section{Experiments}

\subsection{Experimental Setup}

\subsubsection{Datasets}

\noindent \textbf{SFT Data.}
To bootstrap the agentic framework and enable subsequent online data collection, we first construct SFT data for the base policy. For the strategy selection stage $\mathcal{F}_{\mathrm{sel}}$, we sample 50K online product requests and construct stage-specific prompts from their product contexts and material packages. We use Qwen3-VL-235B-A22B-Thinking~\cite{bai2025qwen3vl} as a larger reasoning VLM teacher to generate strategy selection trajectories, and we keep only trajectories that pass output format validation and strategy constraint validation. For the draft generation stage $\mathcal{F}_{\mathrm{draft}}$, we collect 100K strategy selection output trajectories and use the same teacher model to generate draft trajectories. The resulting SFT data is used to train a Qwen3-VL-8B-Thinking model~\cite{bai2025qwen3vl}.

\noindent \textbf{Performance-Based Reward Data.}
For performance-based reward modeling, we collect 1M video pairs over a one-month window of online logs through the impression-balanced delivery pipeline. We sample across time windows to reduce delivery period bias. Because impression-balanced delivery gives a group of videos similar impression opportunities and reduces advertising system bias, these collected pairs provide cleaner labels for pairwise learning. We reserve 10K pairs from the 1M impression-balanced samples as the validation set.

\noindent \textbf{Rubric-Based Reward Data.}
For rubric-based reward modeling, we collect human annotations for video pairs from the \textsc{AgenticGen} framework. Each pair is independently labeled by three human annotators. We retain 10K video pairs for which all three annotators give consistent pairwise labels and split them 9:1 into training and validation sets. We further apply reject sampling with Qwen3-Omni-30B-A3B-Thinking~\cite{xu2025qwen3omni} to construct rubric reward SFT data, so the rubric judge learns to align with the expert criteria before being used as a reward model.

\noindent \textbf{RL Data.}
For reinforcement learning, we also rely on the impression-balanced delivery pipeline to collect fairly comparable pairwise data. DPO uses 50K impression-balanced strategy selection pairs and 50K impression-balanced draft generation pairs, each split 9:1 into training and validation sets, leaving 5K validation pairs for each stage. GRPO uses the input prompts from the same impression-balanced data, because on-policy rollout optimization samples fresh trajectories from the current policy. Therefore, GRPO only requires the input prompts rather than fixed response pairs.

\vspace{-2mm}
\subsubsection{Implementation Details}

\noindent \textbf{Reward Models.}
Both the performance and rubric reward models are implemented with Qwen2.5-Omni-7B~\cite{xu2025qwen25omni}. We train the performance reward model with a learning rate of $5\times10^{-6}$ and the rubric reward model with a learning rate of $1\times10^{-5}$. Both reward models use feature groups including video frames, audio and speech features, dense captions, structured storylines, and ad-specific features. For audio features, we enable \texttt{use\_audio\_in\_video} for all samples to support interleaved audio-video encoding. The performance reward model scores each video with pointwise inputs and is trained with the pairwise Bradley-Terry loss. The rubric reward model receives a pair of videos and outputs the reasoning process and the final preference result.

\noindent \textbf{RL Training.}
We run the DPO and GRPO training phases sequentially, initializing DPO from the SFT policy and GRPO from the DPO policy. For DPO, we set the KL coefficient $\beta=0.1$, use a learning rate of $5\times10^{-6}$, and set the NLL loss weight $\lambda_{\mathrm{nll}}=0.2$. For GRPO, we use a learning rate of $1\times10^{-6}$, a rollout group size of $G=8$, a clipping parameter $\epsilon=0.2$, and a KL coefficient $\beta=0.001$. In the strategy selection stage, we set $\lambda_{\mathrm{sel}}=0.5$ to balance global and local strategy priors. In the draft generation stage, we set $\lambda_{\mathrm{draft}}=0.6$ to fuse performance and rubric scores.

\subsection{Offline Evaluation}

\subsubsection{Evaluation of Performance-Based Reward Model.}

We first use the impression-balanced (IB) validation set to evaluate whether the pairwise reward formulation is more suitable than pointwise prediction for learning from online feedback. The pointwise formulation predicts an absolute positive or negative label for each individual sample. The pairwise Bradley-Terry model instead compares two videos under the same product context, making the learning target more stable and closer to the relative video preference needed by \textsc{AgenticGen}. Table~\ref{tab:rm-pointwise-pairwise} shows that pairwise training reaches 60.85\% accuracy on the IB validation set, exceeding pointwise prediction by 7.93\%. This improvement supports the effectiveness of converting noisy online feedback into product-relative pairwise preference labels. We also note that the absolute accuracy of the pairwise reward model remains moderate. Our analysis suggests that this is partly because the reward model can not observe rich cross features used by production CTR models, such as user behavior sequences and user-ad interaction features. Nevertheless, the learned reward model still captures meaningful online preferences and provides useful guidance for downstream RL optimization.

\begin{table}[t]
  \caption{Comparison of pointwise and pairwise reward modeling on the impression-balanced (IB) validation set.}
  \label{tab:rm-pointwise-pairwise}
  \centering
  \begin{tabular}{lcc}
    \toprule
    Method & IB Acc. $\uparrow$ & Gain \\
    \midrule
    Pointwise & 52.92 & - \\
    Pairwise & \textbf{60.85} & +7.93 \\
    \bottomrule
  \end{tabular}
\end{table}

We then examine the contribution of multimodal features to reward accuracy. As shown in Table~\ref{tab:rm-feature-scaling}, accuracy rises from 56.55\% with video features alone to 60.85\% after all feature groups are added. Audio provides the largest gain of 1.88\%. Background music conveys rhythm and emotion, and we observe that synchronization between musical beats and visual edits is strongly associated with online business performance. ASR further exposes spoken narrative content that is not fully available from video frames. Storyline features add 0.93\% by complementing ASR with explicit video-level narrative structure. Finally, ad-specific features contribute another 1.49\% by making hooks, selling points, and CTA cues easier to compare, leading to a clear improvement.

\begin{table}[t]
  \caption{Feature scaling for the performance-based reward model on the impression-balanced (IB) validation set.}
  \label{tab:rm-feature-scaling}
  \centering
  \begin{tabular}{lcc}
    \toprule
    Input features & IB Acc. $\uparrow$ & Gain \\
    \midrule
    Video features & 56.55 & - \\
    + Audio features & 58.43 & +1.88 \\
    + Storyline features & 59.36 & +0.93 \\
    + Ad-specific features & \textbf{60.85} & +1.49 \\
    \bottomrule
  \end{tabular}
\end{table}

\subsubsection{Evaluation of Rubric-Based Reward Model.}

We evaluate the SFT rubric-based reward model on the human labeled (HL) and impression-balanced (IB) validation sets. The baseline model uses the off-the-shelf Qwen2.5-Omni-7B without task-specific fine-tuning, whereas the SFT model is trained on the rubric reward data. The HL set measures the SFT model alignment with human quality standards, while the IB set measures how closely rubric-based model aligns with online business preferences.

Table~\ref{tab:rubric-eval} shows that SFT increases HL accuracy from 55.40\% to 69.50\% and IB accuracy from 49.20\% to 53.30\%. The substantial gain on the HL set is consistent with the training objective, because the SFT data directly align the model with human quality judgments. Such alignment helps the reward model favor videos that maintain a platform-native style and attractiveness while avoiding common quality defects. The smaller but positive gain on the IB set indicates that human quality standards are related to online business preferences but do not fully determine them. We hypothesize that many evaluated videos already meet a basic quality threshold, after which online performance is further differentiated by the internal factors of advertising system and the heterogeneous audience preferences. This difference between strong human alignment and weaker online transfer supports using the rubric reward as a complementary signal to the performance reward in downstream draft generation optimization.

\begin{table}[t]
  \caption{Performance of the rubric-based reward model on the human labeled (HL) and impression-balanced (IB) validation sets.}
  \label{tab:rubric-eval}
  \centering
  \begin{tabular}{lcc}
    \toprule
    Method & HL Acc. $\uparrow$ & IB Acc. $\uparrow$ \\
    \midrule
    Baseline & 55.40 & 49.20 \\
    SFT & \textbf{69.50} & \textbf{53.30} \\
    \bottomrule
  \end{tabular}
\end{table}

\subsubsection{Evaluation of DPO Optimization.}

The SFT policy learns to produce strategies and drafts in valid formats, but it is not directly optimized to generate responses with strong online business performance. DPO addresses this gap through impression-balanced preference training, in which the preferred response is associated with stronger business performance under controlled delivery. Table~\ref{tab:dpo-eval} compares the DPO and SFT policies by how consistently they align with this preference ordering. Under the DPO formulation, the policy/reference log probability ratio acts as an implicit reward~\cite{rafailov2023direct}. A pair is correctly ranked when the preferred response receives a larger implicit reward than its dispreferred counterpart, and aggregating these comparisons yields preference accuracy~\cite{lambert2025rewardbench,chen2025bootstrapping}. For the SFT baseline, the corresponding ordering is derived from sequence likelihood. Appendix~\ref{app:preference-accuracy} formalizes both scoring rules and the resulting accuracy metric.

Table~\ref{tab:dpo-eval} reports preference accuracy for the strategy selection and draft generation stages. DPO increases strategy selection accuracy from 49.72\% to 56.48\% and draft generation accuracy from 50.64\% to 58.34\%. The average accuracy rises from 50.18\% to 57.41\%. These consistent gains show that impression-balanced preference training provides effective supervision for both agentic reasoning stages.

\begin{table}[t]
  \caption{Performance of DPO for strategy selection and draft generation on the impression-balanced (IB) validation sets.}
  \label{tab:dpo-eval}
  \centering
  \begin{tabular}{lccc}
    \toprule
    Method & Selection Acc. $\uparrow$ & Draft Acc. $\uparrow$ & Avg. Acc. $\uparrow$ \\
    \midrule
    SFT & 49.72 & 50.64 & 50.18 \\
    DPO & \textbf{56.48} & \textbf{58.34} & \textbf{57.41} \\
    \bottomrule
  \end{tabular}
\end{table}

\subsubsection{Ablation of GRPO Rewards for Draft Generation.}

GRPO for draft generation combines two complementary signals: a performance-based reward and a rubric-based reward. Table~\ref{tab:grpo-eval} ablates their individual contributions and their fusion. Unlike DPO, GRPO does not yield a policy/reference implicit reward from which preference accuracy can be computed. We therefore evaluate each optimized policy directly at the level of generated videos using reward models. For every prompt in the validation set, we execute and render the response from the evaluated policy and compare the resulting video with the video generated by the SFT policy for the same prompt. Pairwise wins assigned by the performance-based and rubric-based reward models yield the corresponding performance and rubric win rates. We compare three GRPO reward variants: performance reward only, rubric reward only, and weighted fusion of both rewards. We also include the result from DPO as a reference.

Table~\ref{tab:grpo-eval} reveals how the reward choice changes the optimized policy. DPO attains a 54.18\% performance win rate but remains near chance on rubric at 49.48\%, yielding an average of 51.83\%. With performance reward alone, GRPO raises the two win rates to 61.04\% and 51.74\%, respectively. With rubric reward alone, GRPO reaches 60.24\% on rubric, but its performance win rate falls to 50.78\%. The single reward variants expose a clear trade-off between business alignment and human standards. Weighted fusion retains nearly all of the performance gain at 60.52\%, improves the rubric win rate to 55.86\%, and achieves the highest average of 58.19\%.

\begin{table}[t]
  \caption{Reward model win rates of the DPO and GRPO reward variants. \textbf{Perf. WR} and \textbf{Rubric WR} denote performance-based and rubric-based reward model win rates, respectively.}
  \label{tab:grpo-eval}
  \centering
  \begin{tabular}{lccc}
    \toprule
    Method & Perf. WR $\uparrow$ & Rubric WR $\uparrow$ & Avg. WR $\uparrow$ \\
    \midrule
    DPO & 54.18\% & 49.48\% & 51.83\% \\
    Performance only & \textbf{61.04\%} & 51.74\% & 56.39\% \\
    Rubric only & 50.78\% & \textbf{60.24\%} & 55.51\% \\
    Weighted fusion & 60.52\% & 55.86\% & \textbf{58.19\%} \\
    \bottomrule
  \end{tabular}
\end{table}

\subsection{Online Evaluation}

To further validate the effectiveness of \textsc{AgenticGen}, we conduct online A/B tests in the TikTok advertising system using the impression-balanced delivery pipeline. Table~\ref{tab:online} reports two successive comparisons. Pre-Agent denotes the non-agentic online pipeline before \textsc{AgenticGen}, in which most configurations are manually set, and different components are orchestrated separately. The first comparison evaluates \textsc{AgenticGen} SFT against Pre-Agent, while the second evaluates \textsc{AgenticGen} RL after DPO and GRPO against the SFT policy. For both comparisons, we report relative lifts in CTR (Click-Through Rate), CVR (Conversion Rate), and Advv (Advertiser Value). Advv measures the total advertiser-side value of attributed conversions, valuing each conversion by its bid:

\begin{equation}
\label{eq:advv}
\begin{aligned}
\mathrm{Advv}
&= \mathrm{Conversion}\times \mathrm{Bid} \\
&= \mathrm{Impression}\times \mathrm{CTR}\times \mathrm{CVR}\times \mathrm{Bid}.
\end{aligned}
\end{equation}

Table~\ref{tab:online} first shows that \textsc{AgenticGen} SFT improves CTR by 3.48\%, CVR by 2.30\%, and Advv by 9.83\% over Pre-Agent. These gains capture the system-level benefit of replacing manual configuration with agent decisions and integrating previously separate components into a unified agentic workflow. After the \textsc{AgenticGen} framework is established, the comparison between SFT and RL isolates the algorithmic gain from DPO and GRPO optimization, thereby providing direct evidence for our central claim that reward-guided agentic RL can further improve online business performance beyond the SFT baseline. Compared with SFT, RL increases CTR by 2.72\%, CVR by 2.63\%, and Advv by 9.61\%. These gains indicate that learning from online business feedback through DPO and GRPO improves the CTR of generated advertising videos. Although the collected feedback is mainly organized around CTR, CVR also increases, suggesting that the optimized policy benefits not only click propensity but also downstream conversion. The joint growth of CTR and CVR further leads to the Advv lift. We note that the Advv gain is larger than the multiplicative lift implied by CTR and CVR alone because Advv weights attributed conversions by bid. Our analysis suggests that higher bid traffic often contains stronger source materials, giving \textsc{AgenticGen} more reliable product materials for strategy selection and draft generation. When source materials are weak, the generation tools have less reliable visual reference for transforming them into more effective advertising videos.

\begin{table}[t]
  \caption{Online A/B evaluation in the TikTok advertising system.}
  \label{tab:online}
  \centering
  \begin{tabular}{lccc}
    \toprule
    Optimization Stage & CTR $\uparrow$ & CVR $\uparrow$ & Advv $\uparrow$ \\
    \midrule
    Pre-Agent $\rightarrow$ SFT & +3.48\% & +2.30\% & +9.83\% \\
    SFT $\rightarrow$ RL & \textbf{+2.72\%} & \textbf{+2.63\%} & \textbf{+9.61\%} \\
    \bottomrule
  \end{tabular}
\end{table}

\section{Conclusion}

In this paper, we introduce \textsc{AgenticGen}, a reward-guided agentic framework that addresses the challenge of optimizing advertising video generation for online business performance. Rather than treating video generation as a single synthesis step, \textsc{AgenticGen} formulates it as product-conditioned reasoning problem and decomposes the process into two trainable reasoning stages for strategy selection and draft generation. The framework learns a performance-based reward model from impression-balanced delivery data and a complementary rubric-based reward model aligned with domain-specific human quality standards, then applies DPO and GRPO to improve both reasoning stages. Online A/B experiments in the TikTok advertising system demonstrate that \textsc{AgenticGen} after DPO and GRPO improves CTR by 2.72\%, CVR by 2.63\%, and Advv by 9.61\% over the SFT baseline, validating the effectiveness of reward-guided agentic video generation at scale.

\clearpage

\bibliographystyle{plainnat}
\bibliography{paper}

\clearpage
\beginappendix
\section{Strategy Definitions}
\label{app:strategy-examples}

The complete strategy catalog contains internal commercially sensitive rules and therefore cannot be fully enumerated. We provide abstract definitions and representative examples below.

\begin{itemize}[leftmargin=4mm]
    \item \textbf{Asset editing.} This family applies lightweight transformations to existing product videos or generated videos while preserving their usable content and overall storyline. Representative operations include replacing background music, adding a short hook, inserting readable overlays or stickers, highlighting selling points, and strengthening the call to action.
    \item \textbf{Reference-guided generation.} This family retrieves and interprets high-performing reference videos as narrative context, then adapts their effective storyline patterns to the current product. Representative outputs include a product-specific storyboard, script, and generation instructions for new clips, which may subsequently be combined with source materials through editing or remixing strategies.
    \item \textbf{Cross-asset remixing.} This family constructs a revised storyline by selecting, ordering, and combining eligible clips from product materials. Representative actions include clip selection, scene ordering, and narration planning. It also includes segment replacement as a special case: clips with weak product presentation, low information density, or insufficient visual appeal may be removed, replaced, or extended with stronger source or newly generated clips.
\end{itemize}

\section{Strategy Selection Reward Details}
\label{app:selection-reward-details}

Let $\mathcal{D}^{\mathrm{sel}}=\{(S_t,b_t)\}_{t=1}^{T}$ denote the historical impression-balanced delivery logs, where $S_t=\{s_{t,k}\}_{k=1}^{K_t}$ is the selected strategy group and $b_t$ is scalar online business feedback, \ie, CTR. To estimate the global performance trend of each strategy $s$, each group record is expanded into $(s,b_t)$ for every $s\in S_t$. This expansion produces $\sum_{t=1}^{T}|S_t|$ single-strategy assignments. We then average the feedback assigned to each distinct strategy across products and campaigns and score a candidate group by the mean of its strategy scores. Algorithm~\ref{alg:global-strategy-reward} summarizes this procedure.

\begin{algorithm}[t]
\caption{Global strategy reward construction}
\label{alg:global-strategy-reward}
\begin{algorithmic}
\STATE \textbf{Input:} Logs $\mathcal{D}^{\mathrm{sel}}=\{(S_t,b_t)\}_{t=1}^{T}$
\STATE \textbf{Input:} Candidate strategy group $S$
\STATE \textbf{Output:} Global reward $R_{\mathrm{glob}}$
\STATE Initialize an empty record list $\mathcal{E}$
\FOR{each $(S_t,b_t)$ \textbf{in} $\mathcal{D}^{\mathrm{sel}}$}
    \FOR{each $s$ \textbf{in} $S_t$}
        \STATE Append $(s,b_t)$ to record list $\mathcal{E}$
    \ENDFOR
\ENDFOR
\FOR{each $s$ appearing in $\mathcal{E}$}
    \STATE $\bar b_s\gets\operatorname{mean}\{b:(s,b)\in\mathcal{E}\}$
\ENDFOR
\STATE $R_{\mathrm{glob}}\gets |S|^{-1}\sum_{s\in S}\bar b_s$
\STATE \textbf{Return:} $R_{\mathrm{glob}}$
\end{algorithmic}
\end{algorithm}

For the local reward, let $\{S_i\}_{i=1}^{N}$ denote the historical delivered strategy groups under the same product request as the current rollout samples. We select the strategy group with the highest feedback as the local winner $S^+$ and score a candidate rollout sample $S$ by its Jaccard overlap with $S^+$. Algorithm~\ref{alg:local-strategy-reward} summarizes the procedure.

\begin{algorithm}[t]
\caption{Local strategy reward construction}
\label{alg:local-strategy-reward}
\begin{algorithmic}
\STATE \textbf{Input:} Delivered strategy groups $\{S_i\}_{i=1}^{N}$ and corresponding feedback $\{b_i\}_{i=1}^{N}$
\STATE \textbf{Input:} Candidate strategy group $S$
\STATE \textbf{Output:} Local reward $R_{\mathrm{loc}}$
\STATE $i^+\gets\arg\max_{i\in\{1,\ldots,N\}} b_i$
\STATE $S^+\gets S_{i^+}$
\STATE $R_{\mathrm{loc}}\gets |S\cap S^+|/|S\cup S^+|$
\STATE \textbf{Return:} $R_{\mathrm{loc}}$
\end{algorithmic}
\end{algorithm}

\section{Preference Accuracy}
\label{app:preference-accuracy}

For each reasoning stage, strategy selection or draft generation, the held-out impression-balanced validation set, denoted by $\mathcal{D}_{\mathrm{val}}^{\mathrm{sel}}$ or $\mathcal{D}_{\mathrm{val}}^{\mathrm{draft}}$, contains triples $(X,Y^+,Y^-)$, where $Y^+$ and $Y^-$ are the policy responses associated with higher and lower online business feedback under the same stage input $X$. The evaluation examines whether a policy score recovers this online preference ordering.

DPO admits an implicit reward through the optimized policy $\pi_\psi$ and its SFT reference policy $\pi_{\mathrm{ref}}$~\cite{rafailov2023direct,lambert2025rewardbench,chen2025bootstrapping}:
\begin{equation}
\label{eq:appendix-dpo-implicit-reward}
r_\psi(X,Y)
=
\beta \log \frac{\pi_\psi(Y \mid X)}{\pi_{\mathrm{ref}}(Y \mid X)}
+ \beta \log Z(X),
\end{equation}
where $Z(X)$ is the input-dependent partition function. Because $Y^+$ and $Y^-$ share the same input, the partition term cancels when their rewards are compared. We therefore compute the DPO preference margin as
\begin{equation}
\label{eq:appendix-dpo-preference-margin}
\begin{aligned}
\Delta_{\mathrm{DPO}}(X,Y^+,Y^-)
&= \beta\Bigg[
\log \frac{\pi_\psi(Y^+ \mid X)}{\pi_{\mathrm{ref}}(Y^+ \mid X)}\\
&\qquad\quad-
\log \frac{\pi_\psi(Y^- \mid X)}{\pi_{\mathrm{ref}}(Y^- \mid X)}
\Bigg].
\end{aligned}
\end{equation}
A positive margin means that the implicit reward ranks the online-preferred response higher.

The SFT baseline has no implicit reward. Its corresponding preference margin is computed directly from sequence likelihood:
\begin{equation}
\label{eq:appendix-sft-preference-margin}
\Delta_{\mathrm{SFT}}(X,Y^+,Y^-)
=
\log \pi_{\mathrm{SFT}}(Y^+ \mid X)
-\log \pi_{\mathrm{SFT}}(Y^- \mid X).
\end{equation}
For both SFT and DPO, we compute stage-specific preference accuracy as
\begin{equation}
\label{eq:appendix-dpo-preference-accuracy}
\mathrm{Acc}
=
\frac{1}{|\mathcal{D}_{\mathrm{val}}|}
\sum_{(X,Y^+,Y^-)\in\mathcal{D}_{\mathrm{val}}}
\mathbb{I}\!\left[\Delta(X,Y^+,Y^-)>0\right],
\end{equation}
where $\mathbb{I}$ is an indicator function that returns $1$ if the condition is true and $0$ otherwise.

\end{document}